# A New Dynamic Muscle Fatigue Model and its Validation

Liang MA[a]   Damien CHABLAT[a]   Fouad BENNIS[a]   Wei ZHANG[b]

[a] Institut de Recherche en Communications et en Cybernetique de Nantes,
UMR CNRS 6597, Ecole Centrale de Nantes
1, rue de la Noë - BP 92 101 - 44321 Nantes CEDEX 03, France

[b] Department of Industrial Engineering, Tsinghua University, Beijing100084, P.R.China

**ABSTRACT**

Musculoskeletal disorder (MSD) is one of the major health problems in physical work especially in manual handling jobs. In several literatures, muscle fatigue is considered to be closely related to MSD, especially for muscle related disorders. In addition to many existing analysis techniques for muscle fatigue assessment and MSD risk analysis, in this paper, a new muscle fatigue model was proposed. The new proposed model reflects the influence of external load, workload history, and individual differences. This model is simple in mathematics and can be easily applied in realtime calculation, such as the application in realtime virtual work simulation and evaluation. The new model was mathematically validated with 24 existing static models by comparing the calculated *MET*s, and qualitatively or quantitatively validated with 3 existing dynamic models. The proposed model shows high or moderate similarities in predicting the *MET*s with all the 24 static models. Validation results with the three dynamic models were also promising. The main limitation of the model is that it still lacks experimental validation for more dynamic situations.

**Relevance to industry**

Muscle fatigue is one of the main reasons causing MSDs in industry, especially for physical work. Correct evaluation of muscle fatigue is necessary to determine work-rest regimens and reduce the risks of MSD.

# 1. INTRODUCTION

Musculoskeletal disorder is defined as injuries and disorders to muscles, nerves, tendons, ligaments, joints, cartilage and spinal discs and it does not include injuries resulting from slips, trips, falls or similar accidents (Maier and Ross-Mota, 2000). From the report of HSE (HSE, 2005) and the report of Washington State Department of Labor and Industries (Safety and Health Assessment and Research for Prevention (SHARP), 2005), over 50% of workers in industry have suffered from musculoskeletal disorders. There are numerous of \risk factors" associating with the work-related MSDs, such as physical workload factors (Burdorf, 1992), psychosocial factors (Bongers et al., 1993) and individual factors (Armstrong et al., 1993). According to the analysis in Occupational Biomechanics (Chaffin and Andrersson, 1999) and in several literatures (Armstrong et al., 1993; Buckle and Devereux, 2002), "physical work requirements and individual factors determine muscle force and length characteristics as a function of time, which in turn determines muscle energy requirements. Muscle energy requirements in turn can lead to fatigue, which then can lead to muscle disorders." Overexertion of muscle force or frequent high muscle load is the main reason for muscle fatigue, and further more, it results in acute muscle fatigue, pain in muscles and the worst functional disability in muscles and other tissues of human body. Hence, it is necessary for ergonomists to find an efficient method to assess physical exposures to muscles and to predict muscle fatigue in work design stage.

In order to assess physical risks to MSDS, several ergonomics tools were developed and most of them were listed, classified and compared in paper (Li and Buckle, 1999). Observational methods for posture analysis, such as Posturegram, Ovako Working Posture Analyzing System (OWAS), Posture Targeting and Quick Exposure Check for work-related musculoskeletal risks (QEC), were developed for analyzing whole body postures. In all these four methods, posture is taken as one of the most important factors to assess the physical exposure. In the former three methods, body posture is categorized into different types with different risk levels according to recorded position. The differences between these methods are the rules to classify the body positions. In QEC method, posture of different body parts is scaled into different exposure levels. In combination with posture, other physical factors like force, repetition and duration of movement, are also taken into consideration to assess physical workload in OWAS and QEC methods. In spite of these general posture analysis tools, some special tools are designed for specific parts of human body. For example, Rapid Upper Limb Assessment (RULA) is designed for assessing the severity of postural loading for upper extremity. This method has the same concept of OWAS, but particularly suited for sedentary jobs (McAtamney and Corlett, 1993). It uses ranking system to rate different postures, different movements and repetition/duration of the task. The similar systems include HAMA (Hand-Arm-Movement Analysis), PLIBEL (method for the identification of musculoskeletal stress factors that may have injurious effects) and so on (Stanton et al., 2004). \In general, these observational methods are mainly posture-based. They are relatively inexpensive to carry out, and the assessments can be made without disruption to the workforce" (Li and Buckle, 1999). Similar to these methods for posture analysis, there is one tool available for fatigue analysis: muscle fatigue analysis (MFA) (Rodgers, 2004). This technique is developed by Rodgers and Williams to characterize the discomfort described by workers on automobile assembly lines and fabrication tasks (Rodgers, 1987). In this method, each body part is scaled into four effort levels according to its working position, and meanwhile the duration of the effort and frequency are both scaled into four effort levels. The combination of the three factors' levels can determine "priority to change" score. The task with high priority score needs to be analyzed and redesigned to reduce the MSD risks (Stanton et al., 2004).

After listing these available methods, physical exposure to MSD can be evaluated with respect to intensity (or magnitude), repetitiveness and duration (Li and Buckle, 1999). While these methods can be used to assess physical jobs, there are still several limitations. At first, even just for lifting job, the evaluation results of five tools (NIOSH lifting index, ACGIH TLV, 3DSSPP, WA L&I, Snook) for the same task were different, and sometimes even contradictory (Russell et al., 2007). That is because these techniques are lack of precision and their reliability of the system is a problem for assessing the physical exposures due to their intermittent recording procedures (Burdorf, 1992). Second, most of the traditional methods have to be carried out onsite; therefore there is no immediate result from the observation. It is time consuming for afterward analysis. Further more, subjective variability can influence the evaluation results even using the same observation methods for the same task (Lamkull et al., 2007). At last, only intermittent posture positions and limited working conditions are considered in these methods, which means that they are suitable for analyzing static working process and they are not less suitable to estimate the MSD risks into details.

Besides these objective posture analysis tools, there are several self-report methods to assess the physical load or body discomfort, such as "body map" (Corlett and Bishop, 1976), rating scales (Borg, 1998), questionnaires or interviews (Wiktorin et al., 1993) and checklists (Corlett, 1995). These tools are also important because ergonomists have to concentrate themselves on the feeling of the workers. Several authors even insist that "If the person tells you that he is loaded and effortful, then he is loaded and effortful whatever the behavioral and performances measures may show" (Li and Buckle, 1999). For muscle fatigue, perceived rating exertion (PRE) and Swedish Occupational Fatigue Inventory (SOFI) based on PRE were developed to rate the workload in practice (Borg, 1998; Ahsberg et al., 1997; Ahsberg and Gamberale, 1998). SOFI consists of five aspects: lack of motivation, sleepiness, physical discomfort, lack of energy, and physical exertion, and it is used to measure fatigue as a perception of either mental or physical character (Ahsberg and Gamberale, 1998). The concept of perceived exertion and the associated methods for measuring fatigue is: "the human sensory system can function as an efficient instrument to evaluate the work load by integrating many peripheral and central signals of strain" (Borg, 2004). These subjective assessments of body strain and discomfort have been the most frequently used form due to the ease of use and apparent face validity. But subjective ratings are prone to many influences. This kind of approach has lower validity (Burdorf and Laan, 1991) and reliability (Wiktorin et al., 1993).

In order to evaluate the human work objectively and quickly, virtual human techniques have been developed to facilitate the ergonomic evaluation, such as Jack (Badler et al., 1993), ErgoMan (Schaub et al., 1997), 3DSSPP (Chaffin,1969), Santos (VSR Research Group, 2004; Vignes, 2004) and so on. They have been used in the fields of automotive, military, aerospace and so on. These human modeling tools are mainly used for visualization to provide information about body posture, reachability and field of view etc (Lamkull et al., 2007). The effort of combining these virtual human tools with existing posture analysis methods have been done. For example, in the literature (Jayaram et al., 2006), a method to link virtual environment (Jack) and quantitative ergonomic analysis tools (RULA) in real time for occupational ergonomics studies was presented, and it allowed that ergonomic evaluation could be carried out in real time in their prototype system. But until today, there is still no muscle fatigue index available in these virtual human tools for dynamic working process. It is necessary to develop the muscle fatigue model and then integrate it into the virtual human software to evaluate the muscle fatigue and analyze the physical work into details.

For objectively predicting muscle fatigue, several muscle fatigue models and fatigue index have been proposed in literatures. In a series of publications (Wexler et al., 1997; Ding et al., 2000a,b, 2003), Wexler and his colleagues have proposed a new muscle fatigue model based on $Ca^{2+}$ cross-bridge mechanism and verified the model with stimulation experiments, but it is mainly based on physiological mechanism and it seems complex for ergonomic application due to its amount of variables. For example, only for quadriceps, there are more than 20 variables to describe the muscle fatigue mechanism. Further more, there are only parameters available for quadriceps, and it is hard to integrate it for full body application. This model was integrated into virtual soldier research (VSR) system to simulate the movement of legs by lifting loads using quadriceps (Vignes, 2004), and the results showed Wexler's Model could predict muscle fatigue correctly, but it still needs to be generalized for the other muscles. Another muscle fatigue model based on force-pH relationship is developed by (Giat et al., 1993). This fatigue model was obtained by curve fitting of the pH level with time t in the course of stimulation and recovery. Taku Komura et al. (1999, 2000) have employed this model in computer graphics to visualize the muscle capacity and then to evaluate the feasibility of the movement. They did not evaluate the muscle fatigue of the whole working process. Meanwhile in this pH muscle fatigue model, although the force generation capacity can be mathematically analyzed, the influences on fatigue from muscle forces are not considered. Rodriguez proposed a half-joint fatigue model in the publications (Rodriguez et al., 2003a,b,c), more exactly a fatigue index, based on mechanical properties of muscle groups. This fatigue model was used to calculate the fatigue at joint level: two half-joints, and the fatigue level is expressed as the actual holding time normalized by maximum holding time of the half-joint. With this model, it is able to apply a posture optimization algorithm to adapt human posture during a working process dynamically when fatigue appears, but it cannot predict individual muscle fatigue due to its half-joint principle because the movement of joint is activated by several muscles. The maximum holding time equation of this model was from static posture analysis and it is mainly suitable for evaluating static postures. In Liu et al. (2002), a dynamic muscle model is proposed based on motor units pattern of muscle, and in this model, two phenomenological parameters were introduced to construct the muscle model to describe the activation, fatigue and recovery process. But there were just parameters available under maximum voluntary contraction situation of right hand which is rare in manual handling work, and further more, there is still no application of this model in ergonomics field.

In order to analyze the physical work in details and predict the physical exposures, especially muscle fatigue, a new method, concerning the overall dynamic working process, should be developed to assess and predict the potential MSD risks, objectively and easily. In this paper, we are going to propose a dynamic muscle fatigue model and a fatigue index with consideration of muscle load history and personal factors. This fatigue model is going to be validated with comparison of the previous static endurance time models and three dynamic muscle fatigue models. At last, we are going to present an application framework of our dynamic fatigue model to evaluate physical work.

## 2. DYNAMIC MUSCLE FATIGUE MODEL

In order to construct the new fatigue model and fatigue index, like in the mentioned ergonomics methods for physical exposures, external load of the muscle with time and the strength capacity of the muscle are involved in our model. These factors can represent the physical risk factors mentioned before: the external load exposed to the muscle with time can cover the information of the intensity (or magnitude), repetitiveness, and duration of force; and the muscle strength capacity can be determined individually and can be treated as personal factor. Thus, the muscle force history (external factor) and maximum voluntary contraction (MVC) (internal factor) are taken to construct our muscle fatigue model. MVC is defined as "the force generated with feedback and encouragement, when the subject believes it is a maximal effort" (Vollestad, 1997). The effect of MVC on endurance time is often used in ergonomic applications to define the worker capabilities or to decide the work-rest rhythm (Garg et al., 2002). In our model, MVC describes the maximum force generation capacity of an individual muscle.

Our new objective fatigue index is trying to evaluate muscle fatigue by describing the human's perception of muscle fatigue. In general, the fatigue evaluation result is a growth function with external load. In the same period, the larger the external load is, the more fatigue people can feel. The same relation is also used in posture analysis methods with higher risk levels for heavier external load. Meanwhile, the fatigue is a growth function with the reciprocal of muscle force capacity. The smaller the capacity, the quicker the muscle gets fatigue. Further more, fatigue is a growth function with the time. The longer a load is applied, the more fatigue people can feel. This is represented in conventional methods as frequency and duration of physical task. If the fatigue is expressed in differential equation, the influence of time can be excluded. The fatigue index is proposed in Eq. (1). The parameters used in the equations are listed and described in Table 1.

$$\frac{dU(t)}{dt} = \frac{MVC}{F_{cem}(t)} \frac{F_{load}(t)}{F_{cem}(t)} \tag{1}$$

Eq. (1) can be explained as follows:

- $F_{cem}$ describes the capacity of the muscle during the contraction progress at a time instant *t*. It falls down during the contraction process because the muscle gets fatigue in a continuous contraction.
- $F_{load}(t)/F_{cem}(t)$ is the relative load at a time instant t which describes the current muscle force normalized the capacity of the muscle at a time instant *t*. This term describes the relation of fatigue index with normalized relative load.
- $MVC/F_{cem}(t)$: reciprocal of muscle force capacity, this term represents the inverse percentage capacity of the tester at a time instant *t* relative to the initial *MVC*. With development of time, this term gets larger while the $F_{cem}(t)$ falls down, and accordingly the increase of fatigue index becomes slower.

Meanwhile, the capacity of muscle (current muscle force capacity) $F_{cem}(t)$ is changing with time due to the external muscle load. The larger the external load, the faster $F_{cem}(t)$ decreases. The differential equation for $F_{cem}$ is proposed in Eq. (2), which is the basic function of the new dynamic fatigue model.

$$\frac{dF_{cem}(t)}{dt} = -\frac{F_{cem}(t)}{MVC} F_{load}(t) \qquad (2)$$

Eq. (2) can be explained by the motor unit activation pattern of muscle. Muscle is made of muscle fibers. Force and movement of muscle are produced by contraction of muscle fibers controlled by nervous-system command (Liu et al., 2002; Vollestad, 1997). The basic functional unit of muscle is motor unit, which consists of a motoneuron and the muscle fibers that it innervates. The motoneurons supply the control signals from central nervous-system (CNS) to the muscle fibers. A muscle consists of many motor units, the number of which varies depending on the size and function of the muscle. Each motor unit has different force generation capability, and different fatigue and recovery properties. Generally, they can be divided into three types: type I is slow twitch motor units with small force generation capability and low conduction velocity, but a very high fatigue resistance; type IIb is of fast-twitch speed, high force capacity, but fast fatigability; type IIa, between type I and type IIb, has a moderate force capacity and moderate fatigue resistance. The sequence of recruitment is in the order of: I→ IIa → IIb (Vollestad, 1997). For a specified muscle, larger $F_{load}$ means more type II motors are involved into the force generation, as a result, the muscle gets fatigue more rapidly, as expressed in Eq. (2). $F_{cem}$ represents the non-fatigue motor units of the muscle. In the progress of force generation, the number of type II motors is getting smaller and smaller due to fatigue, while the number of the type I motor units remains almost the same due to its high fatigue resistance, the decrease of $F_{cem}$ with the time is getting slower, as expressed in Eq. (2) by term $F_{cem}(t)/MVC$.

The integration result of Eq. (2) is Eq. (3).

$$F_{cem}(t) = MVC e^{\int_0^t -k \frac{F_{load}(u)}{MVC} du} \tag{3}$$

Assume that $F(t)$ is:

$$F(t) = \int_0^t \frac{F_{load}(u)}{MVC} du \tag{4}$$

$MVC$ is a constant value of a muscle or a muscle group for an individual person during a certain period, so we can change Eq. (3) into Eq. (5). If external load $F_{load}$ is constant, assign $C = F_{load}/MVC$, then $F(t) = Ct$, and Eq. (3) can be further simplified into Eq. 5. This constant case can occur during static posture and static load.

$$\frac{F_{cem}(t)}{MVC} = e^{-kF(t)} = e^{-kCt} \tag{5}$$

The subjective perception is a function below, which is closely related to $MVC$ and $F_{load}(t)$. $MVC$ can represent the personal factors (Chaffin and Andrersson, 1999), and $F_{load}(t)$ is the force exerted on the muscle along the time and it reflects the influences of external loads Eq. (6).

$$U(t) = \frac{1}{2k} e^{2kF(t)} - \frac{1}{2k} e^{2kF(0)} \tag{6}$$

In this model, personal factors and external load history are considered to evaluate the muscle fatigue. It can be easily used and integrated into simulation software for real time evaluation especially for dynamic working processes. This model needs to be mathematically validated and ergonomic experimentally validated.

## 3. STATIC VALIDATION

### 3.1 Validation Result

Our dynamic muscle fatigue model is based on the hypothesis on the reduction of the maximum exertable force capacity of muscle. It should be able to describe the most special case: static situations. In static posture analysis, there is no model to describe the reduction of the muscle capacity related to muscle force, but there are several models about maximum endurance time (*MET*) which is a measurement related to static muscular work. *MET* represents the maximum time during which a static load can be maintained (Elahrache et al., 2006). The *MET* is most often calculated in relation to the percentage of the voluntary maximum contraction (*%MVC*) or to the relative force ($f_{MVC}$ = *%MVC/100*) required by the task. These models, cited from (Elahrache et al., 2006), are listed in Table 2.

In our dynamic model, suppose that $F_{load}(t)$ is constant, then it represents the static situation. *MET* is the duration in which $F_{cem}$ falls down to the current $F_{load}$. Thus, *MET* can be figured out in Eq. (7) and (8).

$$F_{cem}(t) = MVC e^{\int_0^t -k \frac{F_{load}(u)}{MVC} du} = F_{load}(t) \qquad (7)$$

$$t = MET = -\frac{\ln \frac{F_{load}(t)}{MVC}}{k \frac{F_{load}(t)}{MVC}} = -\frac{\ln(f_{MVC})}{k f_{MVC}} \qquad (8)$$

In order to analyze the relationship between *MET* obtained from our dynamic model and the other models, two correlation coefficients are calculated. One is Pearson's correlation r in Eq. (9) and the other one is intraclass correlation *ICC* in Eq. (10). *r* indicates the linear relationship between two random variables and *ICC* can represent the similarity between two random variables. The closer r is to 1, the more the two models are linearly related. The closer *ICC* is to 1, the more similar the models are. $MS_{between}$ is the mean square between different *MET* values in different $f_{MVC}$ values; $MS_{within}$ is the mean square within *MET* values in different models at the same $f_{MVC}$ level. *p* is the number of models in the comparison. In our case, we compare the other models with our dynamic model one by one, thus *p* equals to 2. The calculation results are shown in Table 2 and Fig. 1 to 8.

$$r = \frac{\sum_n (A_n - \bar{A})(B_n - \bar{B})}{\sqrt{\sum_n (A_n - \bar{A})^2 \sum_n (B_n - \bar{B})^2}} \quad (9)$$

$$ICC = \frac{MS_{between} - MS_{within}}{MS_{between} + (p-1)MS_{within}} \quad (10)$$

Since Huijgens' model (General model) was developed using data from Rohmert (General model), only Rohmert general model is drawn in Fig. 1 and Fig. 2. And Sjogaard's model was constructed using data from Hagberg (Elbow model) and Rohmert (General model), Sjogaard's model is excluded in Fig. 1 and Fig. 2.

### 3.2 Discussion

From the comparison results of the static validation, it is obvious that *MET* model derived from our dynamic model has a great linear correlation with the other experimental static endurance models, and almost all the Pearson's correlation r are above 0.97. Despite the high linear correlation, there are still large differences between the dynamic model and other *MET* models. These differences include mainly the following influencing factors:

- Experiment methods and model construction: In order to measure *MET*, several tools, such as subjective scales and EMG, are involved. The subjective scales and the variability of participants can bring significant differences into *MET* result (Elahrache et al., 2006). Further more, the *MET* models are constructed using different mathematical models, mainly power function and negative exponential function. But the negative exponential function can not describe the two asymptotic tendencies: a tendency towards infinity for low *%MVC* and a tendency towards zero for values bordering on 100%, *%MVC*. In fact, all the parameters of the other *MET* models are fitted from experimental data, and due to the limitations of sampling amount, the *MET* models can be quite different, especially for the two extreme *%MVC*s.

- Muscle group and posture variability: *MET* models for different muscle groups are different, and the statistics results showed that there is a significant difference between the *MET* values for the back/hip and the *MET* values for the upper limbs, for the same *%MVC* value (Elahrache et al., 2006). In addition, *MET* models are mathematically different even for the same muscle group under different postures. Different muscle group has different anatomical structure and different complexity. In the same muscle group, the involvement of muscle elements can be changed significantly, which can also explain the significance between different postures. In the literature (Garg et al., 2002), the influences of shoulder postures were discussed. It indicated that different posture would produce different moments and loads on the same muscle group, thus it would cause different *MET* curves.

- Interindividual variability: from the figures, it is obvious that the differences of *MET* values are greater for low *%MVC* than those for high *%MVC*. The significant interindividual differences in *MET* (Elahrache et al., 2006) can cause the differences.

From *ICC* column, it shows high similarity between the dynamic model and several *MET* models: for elbow and hand models, 5 out of 7 are higher than 0.90; for general model, 3 out of 5 (Huijgens' and Sjogaard's model are not counted) *ICC* values are higher than 0.85. But it also shows moderate or low similarity with the other *MET* models. For example, with the back/hip models, *ICC* varies from -0.057 to 0.9447. The explanation is: *ICC* correlation is significantly influenced by the complexity of the anatomical structure. In shoulder and back/hip of human body, the anatomical structure is in a much more complex way than in the elbows and hands. For this reason, in these experimental models, the measurement of *MVC* and *MET* is an overall performance of the complex muscle group, but not *MET* of an individual muscle or simple muscle group. Meanwhile, even for one muscle group, in Fig. 7, the differences between the experimental models for hip/back are greater than *MET* models for other muscle groups, e.g. in elbow models (Fig. 5). It can also be explained by the complexity of the anatomical structure. In different working conditions (for example, different postures), the engagement of the muscles in the task in the hip/back of human body and the contraction of muscle are different as well, which can further influence the experimental result.

In conclusion, the dynamic model is validated by comparing with 24 static *MET* models. The validated result shows high similarity with many of the static *MET* models, while moderate similarity with a few static *MET* models, possibly due to complex muscle structure, mathematical function limitation, and measurement condition.

## 4. DYNAMIC VALIDATION
### 4.1 Validation result

Static validation results have shown promising result for general static load and even for some specific body parts. But static procedures are still quite different from dynamic situations, thus our dynamic model need to be examined with the other dynamic models. For this objective, the following part is going to verify our dynamic model through comparison with some existing muscle fatigue models, quantitatively or qualitatively.

In the literature (Freund and Takala, 2002), a muscle fatigue model was proposed and integrated into a dynamic model of forearm. In this model, the muscle was treated like a kind of reservoir, and force production capacity $S^0$ reduces with the time that the muscle is contracted. $S^0$ varies between 0 and the upper limits of the muscle force $S^1$. In this model, the recovery and decay rates depend on $S^1$-$S^0$ and muscle force $S$ (Eq. (11)). The constants $α$ and $β$ were obtained by fitting the solution using experimental results from static endurance time test. In this model, muscle force is taken into consideration as a factor causing muscle fatigue, and further more, muscle force production capacity $S^0$ was proposed just like in our dynamic model $F_{cem}$ to describe the capacity of the muscle after performing certain task. But in this model, the force production capacity and the muscle force are decoupled with each other, which is different in our model. In this model, the same concept was employed to describe the fatigue mechanism of muscle.

$$\frac{dS^0}{dt} = \beta(S^l - S^0) - \beta S \tag{11}$$

Wexler's dynamic muscle fatigue model based on Ca$^{2+}$ cross-bridge mechanism can also verify our dynamic model qualitatively. This model can be used to predict the muscle force fatigue under different stimulation frequencies. The stimulation frequency is to simulate the control commands of CNS for muscle contraction. The higher frequency stimulation, the more muscle motors are activated to generate contraction force, so it represents higher muscle load. From Fig. 9, it is clear that the higher the stimulation frequency is, the larger the force can be generated by the muscle. The larger the peak force (higher frequency) is, the faster the curve declines, and the quicker the muscle gets fatigue. This trend is similarly represented in our dynamic model by Eq. (2).

Though qualitatively verified, it is impossible to verify our dynamic model with Wexler's model quantitatively due to the way in which Wexler's model was obtained. Wexler's model is experimentally validated in stimulation trials, and all parameters were calculated from external stimulation experiment. But the muscle simulated in external manner, the motor recruitment mechanism could be different from that controlled by CNS. In external manner, all the motor units of muscle are stimulated simultaneously, creating a larger force than voluntary contraction and get fatigue more rapidly (Vignes, 2004).

In the literature (Liu et al., 2002), the dynamic model of muscle activation, fatigue and recovery was proposed. This model is based on biophysical mechanisms: motor units pattern in 2. The generated force is proportional to the activated motor units in the muscle. The brain effort B, fatigue property F and recovery property R of muscle can decide the number of activated motor units. The relationship is expressed by Eq. (12). The parameters in this equation are explained in Table 3.

$$\begin{aligned}\frac{dM_A}{dt} &= BM_{uc} - FM_A + RM_F \\ \frac{dM_F}{dt} &= FM_A - RM_F \\ M_{uc} &= M_0 - M_A - M_F\end{aligned} \tag{12}$$

When $t = 0$ under the initial conditions of $M_A = 0$, $M_F = 0$, $M_{uc} = M_0$, we can have Eq. (13) and (14).

$$\frac{M_A(t)}{M_0} = \frac{\gamma}{1+\gamma} + \frac{\beta}{(1+\gamma)(\beta-1-\gamma)} e^{-(1+\gamma)Ft} - \frac{\beta-\gamma}{\beta-1-\gamma} e^{-\beta Ft} \quad (13)$$

$$\frac{M_{uc}(t)}{M_0} = e^{-\beta Ft} \quad (14)$$

In the new dynamic fatigue model, we assume that there is no recovery during physical work, and the workers are trying their best to finish the task which means the brain effort is infinitely high. In this assumption, we set $\gamma=0$ and $\beta \rightarrow$ infinity, then Eq. (15) represents the motor units which are not fatigued in the muscle. The activated motors $M_A(t)/M_0$ and the motors in rest $M_{uc}(t)/M_0$ represent the muscle force capacity. We can simplify the sum of Eq. (13) and (14) to Eq. (15) which do have the same form of our dynamic model Eq. (5).

$$\frac{M_A(t)}{M_0} + \frac{M_{uc}(t)}{M_0} = \frac{\gamma}{1+\gamma} + \frac{\beta}{(1+\gamma)(\beta-1-\gamma)} e^{-(1+\gamma)Ft} + \frac{1}{\beta-1-\gamma} e^{-\beta Ft} = e^{-Ft} \quad (15)$$

This fatigue model has been experimentally verified in (Liu et al., 2002). In the experiment, each subject performed an *MVC* of the right hand by gripping a hand grip device for 3 min. And the fitting curve from the experimental result has almost the same curve of our model in *MVC* condition (Fig. 10). In this model, *F* and *R* are assumed to be constant for an individual under *MVC* working conditions. There is no experimental result for *F* and *R* under the other load situations, thus this muscle fatigue model can only verify our model in *MVC* condition.

**4.2 Discussion**

Through dynamic validation, our dynamic model is either qualitatively or quantitatively verified with the other three existing muscle fatigue models. The fatigue model for the forearm used the same conception like in our fatigue model: the muscle force capacity is related to muscle force with time. Wexler's model based on $Ca^{2+}$ cross-bridge shows the reduction of the muscle force during the time under different stimulation frequencies, the reduction of the muscle capacity shows the same trend like in our muscle fatigue model. With comparison to the active motor model, the muscle force can be expressed in the same form under extremity situation. But in the active motor model, only parameters are available for *MVC* contraction case. The active motor does not supply further validation for other load situations.

## 5. APPLICATION IN DIGITAL WORK EVALUATION FRAMEWORK

After mathematical validation of the proposed dynamic model, a virtual reality framework is going to be constructed to apply the fatigue index in virtual environment for digital work evaluation.

The function structure of the framework is shown in Fig. 11. From Section 2, necessary information for dynamic manual handling jobs evaluation consists of posture, external load, repetition, and duration of load. Motion capture techniques can be applied to collect the motion information. The rest information can be analyzed from haptic interfaces or force feedback systems. All the information, such as motion information, force history, and interaction events, is further processed into Objective Work Evaluation System (OWES). The output of the framework is evaluation results of the physical work, and one of these output results is muscle fatigue evaluation.

The prototype of this framework has been constructed for ergonomics studies to measure the performance in two-dimensional panel control (Wang et al., 2006). The new dynamic fatigue index is going to be integrated into the prototype system to evaluate fatigue of physical works.

## 6. CONCLUSIONS

In this paper, a new muscle fatigue model was proposed. Based on this model, a new fatigue index was proposed. The new model was mathematically validated with 24 existing static models by comparing the calculated *MET*s and qualitatively or quantitatively validated with 3 existing dynamic models. The proposed model shows high or moderate similarities in predicting the *MET*s with all the 24 static models. Validation with the three dynamic models was also promising. The new proposed model reflects the influence of external load, workload history, and individual differences. It is simple in mathematics and can be easily applied in realtime calculation, such as the application in virtual work simulation and evaluation.

The main limitation of the model is that it still lacks experimental validation for more dynamic situations.

## ACKNOWLEDGEMENT

This research was supported by the EADS and by the Region des Pays de la Loire (France) in the context of collaboration between the Ecole Centrale de Nantes (Nantes, France) and Tsinghua University (Beijing, P.R.China).

# TABLES

Table 1
Parameters in Dynamic Fatigue Model

| Item | Unit | Description |
|---|---|---|
| $MVC$ | $N$ | Maximum voluntary contraction, maximum capacity of muscle |
| $F_{cem}(t)$ | $N$ | Current exertable maximum force, current capacity of muscle |
| $F_{load}(t)$ | $N$ | External load of muscle, the force which the muscle needs to generate |
| $k$ | $min^{-1}$ | Constant value, 1 |
| $U$ | $min$ | Fatigue Index |
| $\%MVC$ | | Percentage of the voluntary maximum contraction |
| $f_{MVC}$ | | $\%MVC/100$ |

Table 2
Static validation results (Elahrache et al., 2006) $r$ and $ICC$ between Eq. (8) and the other existing MET models in literatures

| Model | Equations in literatures | r | ICC |
|---|---|---|---|
| **General models** | | | |
| Rohmert | $MET = -1.5 + \frac{2.1}{f_{MVC}} - \frac{0.6}{f_{MVC}^2} + \frac{0.1}{f_{MVC}^3}$ | 0.9937 | 0.8820 |
| Monod and Scherrer | $MET = 0.4167 \left(f_{MVC} - 0.14\right)^{-2.4}$ | 0.8529 | 0.6474 |
| Huijgens | $MET = 0.865 \left[\frac{1-f_{MVC}}{f_{MVC}-0.15}\right]^{-2.4}$ | 0.9964 | 0.8800 |
| Sato et al. | $MET = 0.3802 \left(f_{MVC} - 0.04\right)^{-1.44}$ | 0.9992 | 0.8512 |
| Manenica | $MET = 14.88 \exp(-4.48 f_{MVC})$ | 0.9927 | 0.9796 |
| Sjogaard | $MET = 0.2997 f_{MVC}^{-2.14}$ | 0.9935 | 0.9917 |
| Rose et al. | $MET = 7.96 \exp(-4.16 f_{MVC})$ | 0.9897 | 0.7080 |
| **Upper limbs models** | | | |
| *Shoulder* | | | |
| Sato et al. | $MET = 0.398 f_{MVC}^{-1.29}$ | 0.9997 | 0.7188 |
| Rohmert et al. | $MET = 0.2955 f_{MVC}^{-1.658}$ | 0.9987 | 0.5626 |
| Mathiassen and Ahsberg | $MET = 40.6092 \exp(-9.7 f_{MVC})$ | 0.9783 | 0.7737 |
| Garg | $MET = 0.5618 f_{MVC}^{-1.7551}$ | 0.9981 | 0.9029 |
| *Elbow* | | | |
| Hagberg | $MET = 0.298 f_{MVC}^{-2.14}$ | 0.9935 | 0.9921 |
| Manenica | $MET = 20.6972 \exp(-4.5 f_{MVC})$ | 0.9929 | 0.9271 |
| Sato et al. | $MET = 0.195 f_{MVC}^{-2.52}$ | 0.9838 | 0.9712 |
| Rohmert et al. | $MET = 0.2285 f_{MVC}^{-1.391}$ | 0.9997 | 0.7189 |
| Rose et al.2000 | $MET = 20.6 \exp(-6.04 f_{MVC})$ | 0.9986 | 0.9594 |
| Rose et al.1992 | $MET = 10.23 \exp(-4.69 f_{MVC})$ | 0.9943 | 0.7843 |
| *Hand* | | | |
| Manenica | $MET = 16.6099 \exp(-4.5 f_{MVC})$ | 0.9929 | 0.9840 |
| **Back/hip models** | | | |
| Manenica (body pull) | $MET = 27.6604 \exp(-4.2 f_{MVC})$ | 0.9901 | 0.6585 |
| Manenica (body torque) | $MET = 12.4286 \exp(-4.3 f_{MVC})$ | 0.9911 | 0.9447 |
| Manenica (back muscles) | $MET = 32.7859 \exp(-4.9 f_{MVC})$ | 0.9957 | 0.7306 |
| Rohmert (posture 3) | $MET = 0.3001 f_{MVC}^{-2.803}$ | 0.9745 | 0.5353 |
| Rohmert (posture 4) | $MET = 1.2301 f_{MVC}^{-1.308}$ | 0.9989 | 0.7041 |
| Rohmert (posture 5) | $MET = 3.2613 f_{MVC}^{-1.256}$ | 0.9984 | -0.057 |

Table 3
Parameters in Active Motor Model

| Item | Unit | Description |
| --- | --- | --- |
| $F$ | $s^{-1}$ | fatigue factor, fatigue rate of motor units |
| $R$ | $s^{-1}$ | recovery factor, recovery rate of motor units |
| $B$ | $s^{-1}$ | brain effort, brain active rate of motor units |
| $M_0$ | | total number of motor units in the muscle |
| $M_A$ | | number of activated motor units in the muscle |
| $M_F$ | | number of fatigued motor units in the muscle |
| $M_{uc}$ | | number of motor units still in the rest |
| $\beta$ | | $B/F$ |
| $\gamma$ | | $R/F$ |

**FIGURES**

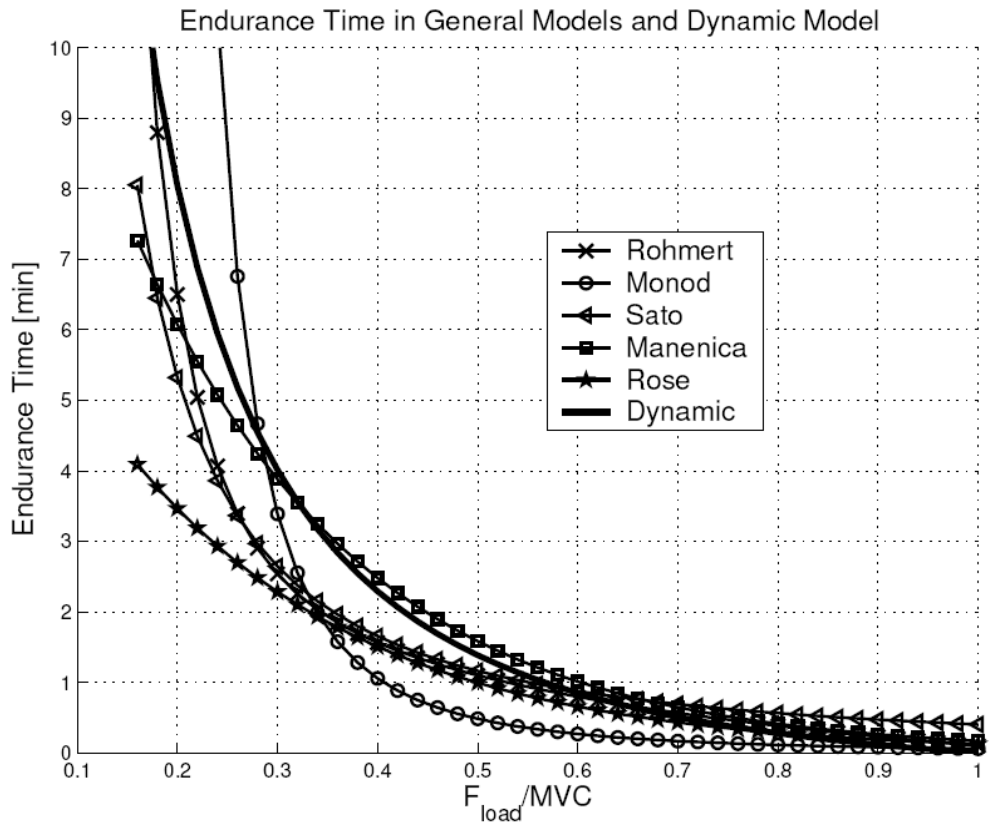

Fig. 1. Endurance time in general models and dynamic model

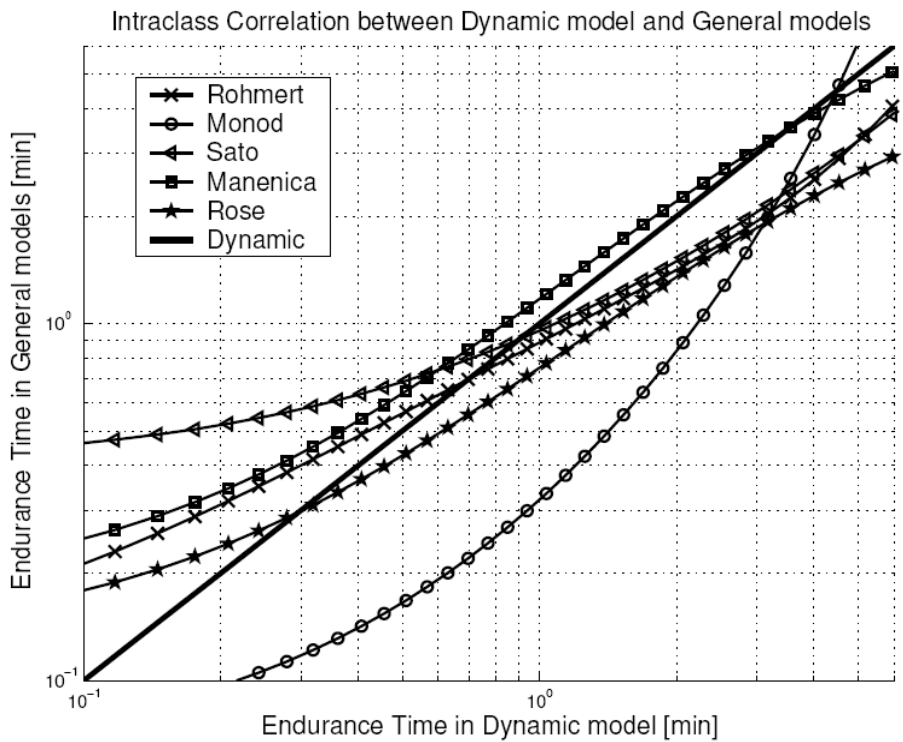

Fig. 2. *ICC* of general models

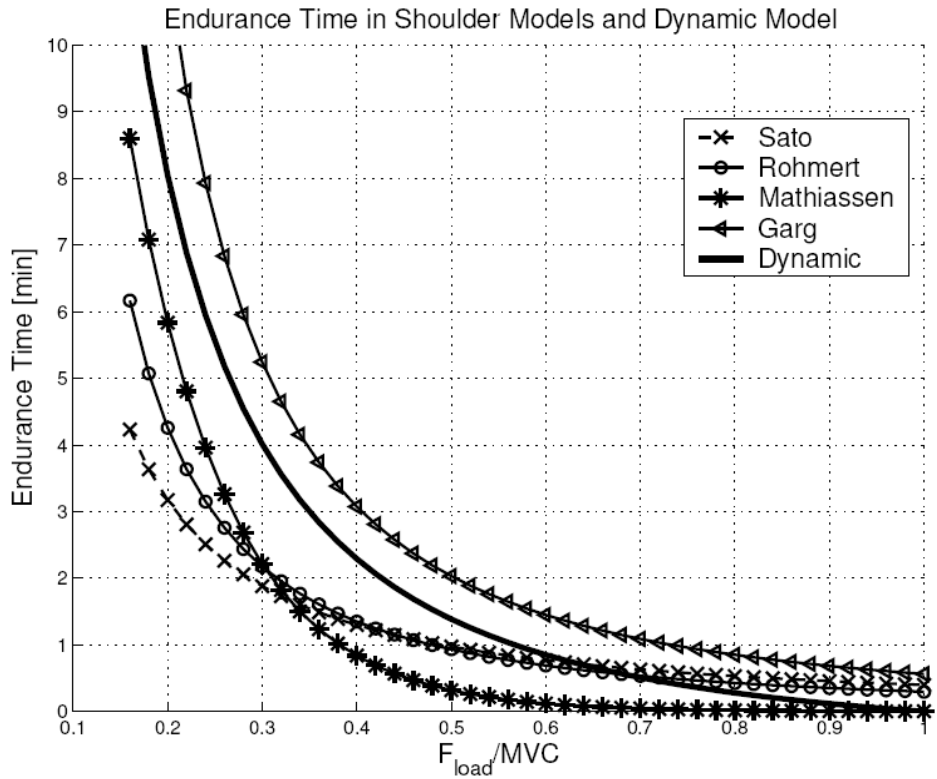

Fig. 3. Endurance time in shoulder endurance models and dynamic model

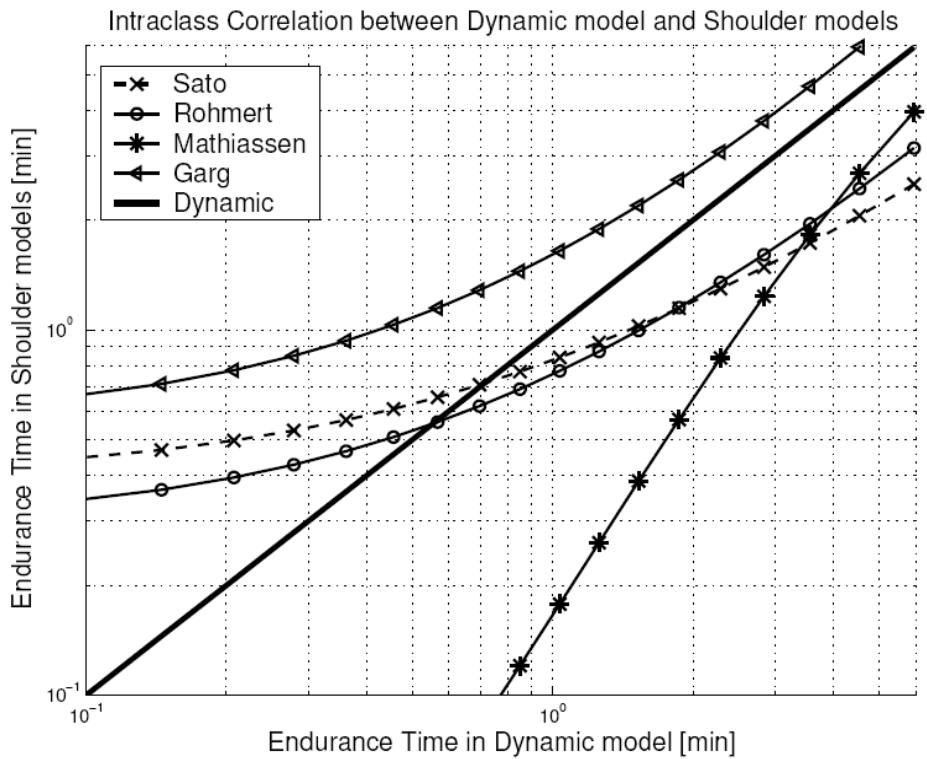

Fig. 4. *ICC* of shoulder endurance models

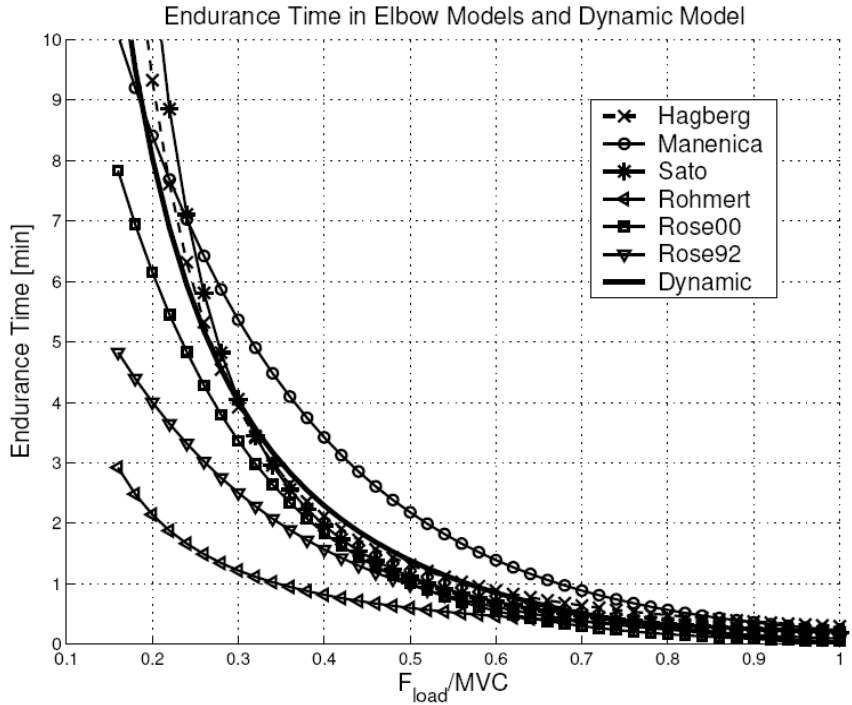

Fig. 5. Endurance time in elbow endurance models and dynamic model

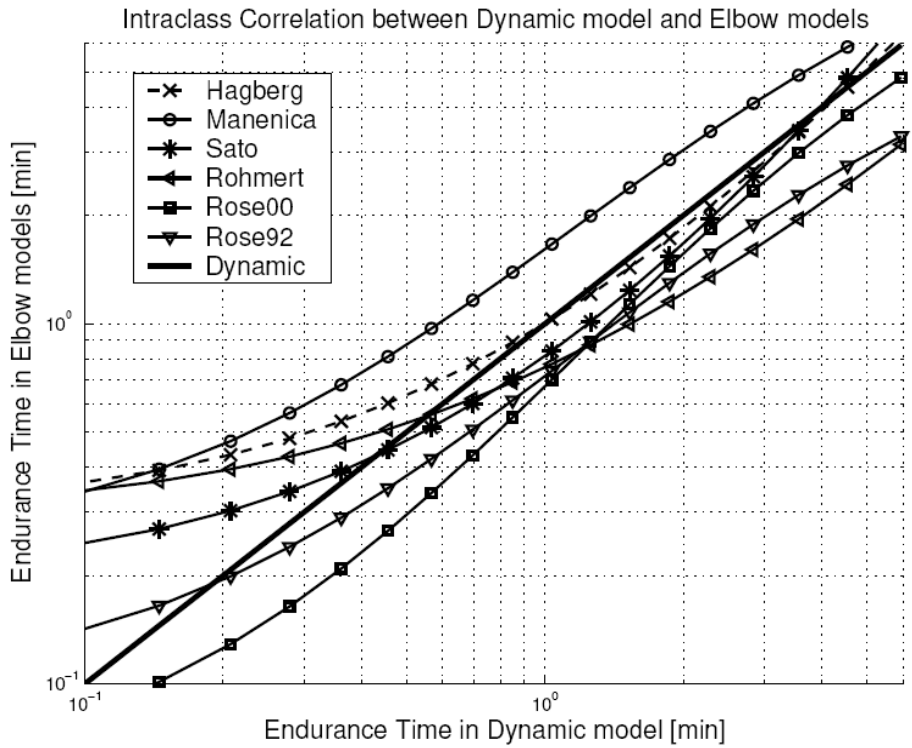

Fig. 6. *ICC* of elbow endurance models

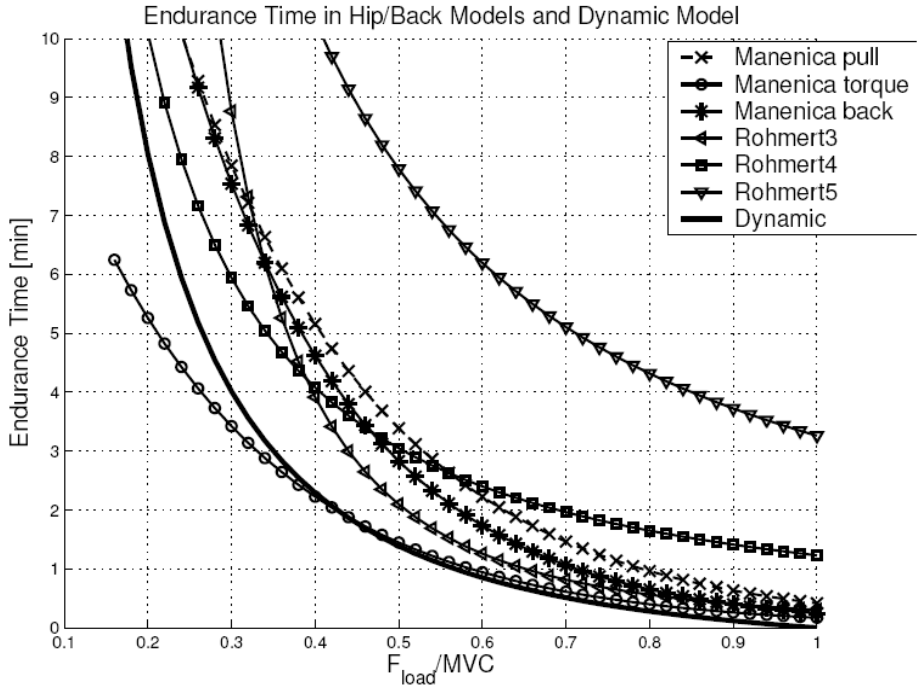

Fig. 7. Endurance time in hip and back models and dynamic model

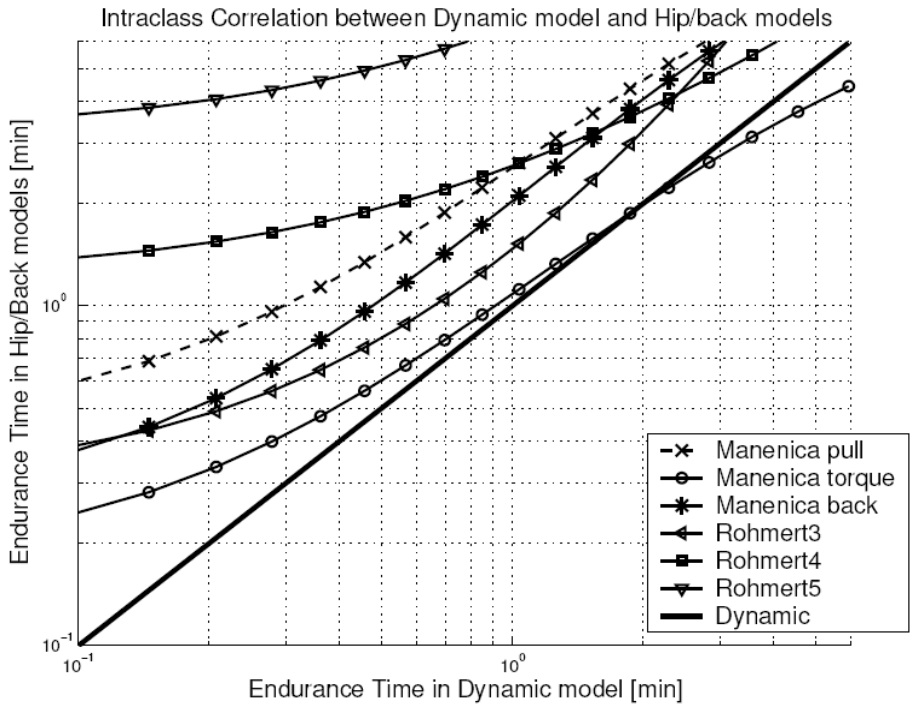

Fig. 8. *ICC* of hip/back models

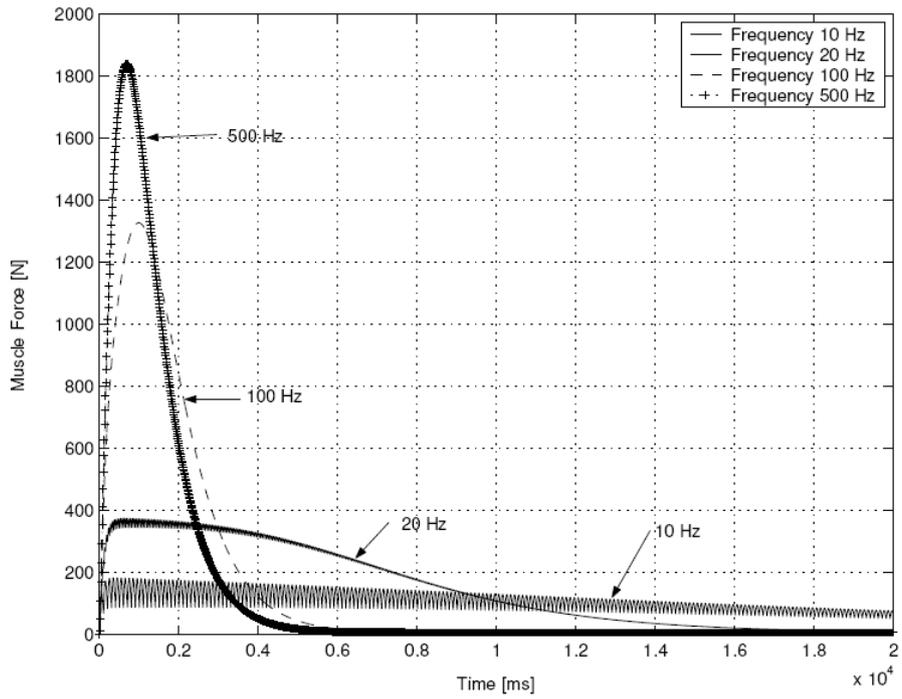

Fig. 9. Maximum exertable force and time relationship in Wexler's Model

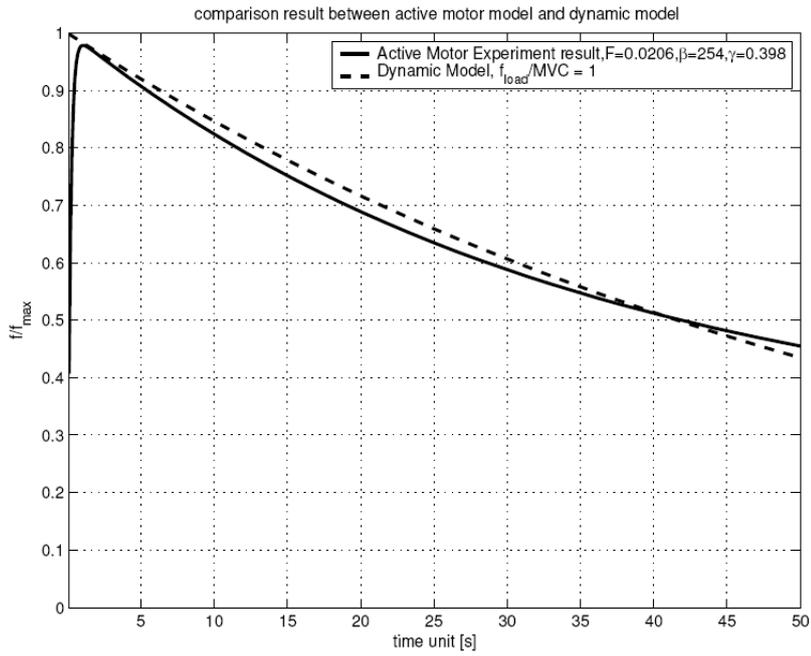

Fig. 10. Comparison between the experimental result of the active motor model and dynamic model in the maximum effort

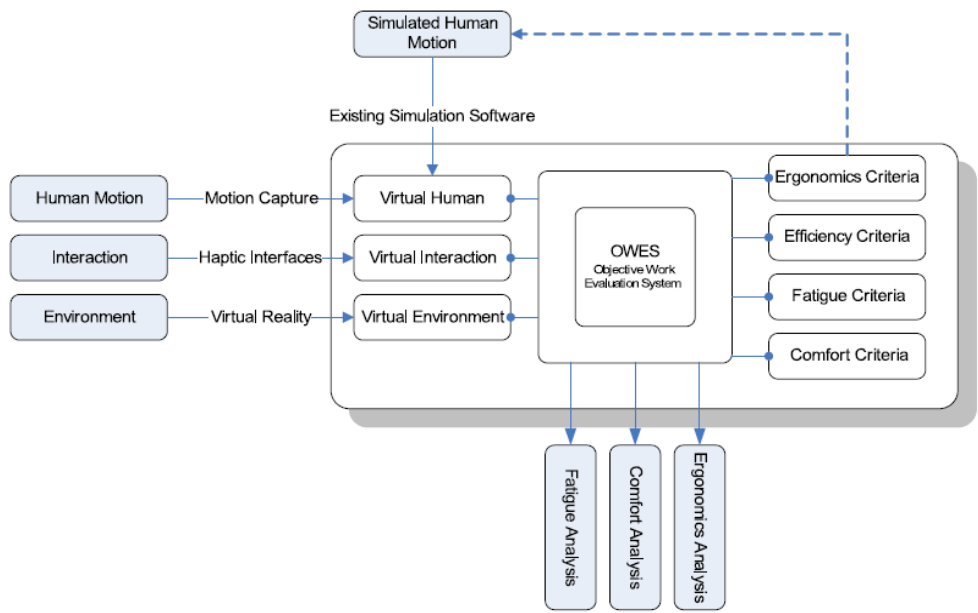

Fig. 11. The framework of the dynamic work evaluation system